\pdfoutput=1

\documentclass[11pt]{article}

\usepackage{EMNLP2023}

\usepackage{amsmath}
\usepackage{multirow}
\usepackage{graphicx}
\usepackage{colortbl}
\usepackage{arydshln} 
\usepackage{color}
\usepackage{makecell}
\usepackage{amsfonts,amssymb}
\usepackage[ruled, vlined, linesnumbered]{algorithm2e}
\usepackage{subfigure}
\usepackage{booktabs}
\usepackage{makecell}

\usepackage{float}
\usepackage{amsfonts}
\usepackage{enumitem}
\usepackage{tikz}
\usepackage{tcolorbox}
\usepackage{siunitx}
\usepackage{amssymb}%
\usepackage{pifont}%

\usepackage{times}
\usepackage{latexsym}

\newcommand{\nop}[1]{}
\newcommand{\old}[1]{}
\newcommand{\new}[1]{#1}

\usepackage[T1]{fontenc}

\usepackage[utf8]{inputenc}

\usepackage{microtype}

\usepackage{inconsolata}
\newcommand\ours{\textsf{OnEFET}\xspace}

\newcommand{\bs}[1]{\boldsymbol{#1}}

\title{Ontology Enrichment for Effective Fine-grained Entity Typing}

\author{Siru Ouyang, Jiaxin Huang, Pranav Pillai, Yunyi Zhang, Yu Zhang, Jiawei Han\\
University of Illinois Urbana-Champaign \\
\normalsize\texttt{\{siruo2, jiaxinh3, ppillai3, yzhan238, yuz9, hanj\}@illinois.edu}\\}

\begin{document}
\maketitle
\begin{abstract}

Fine-grained entity typing (FET) is the task of identifying specific entity types at a fine-grained level for entity mentions based on their contextual information.   
Conventional methods for FET require extensive human annotation, which is time-consuming and costly. Recent studies have been developing weakly supervised or zero-shot approaches.  
We study the setting of zero-shot FET where only 
an ontology is provided.
However, most existing ontology structures lack rich supporting information and even contain ambiguous relations, making them ineffective in guiding FET.
Recently developed language models, though promising in various few-shot and zero-shot NLP tasks, may face challenges in zero-shot FET due to their lack of interaction with task-specific ontology.
In this study, we propose \ours, where we (1) enrich each node in the ontology structure with two types of extra information: \textit{instance} information for training sample augmentation and \textit{topic} information to relate types to contexts,
and (2) develop a coarse-to-fine typing algorithm that exploits the enriched information by training an entailment model with contrasting topics and instance-based augmented training samples. 
Our experiments show that \ours achieves high-quality fine-grained entity typing without human annotation, outperforming existing zero-shot methods by a large margin and rivaling supervised methods.
\end{abstract}

\section{Introduction}

The goal of fine-grained entity typing (FET) is to determine the fine-grained types of entity mentions based on their
contexts. 
This is a crucial task for various text mining and NLP applications, such as entity linking~\cite{chen2020improving}, text classification~\cite{hu2019heterogeneous}, and scientific literature mining~\cite{zhong-etal-2023-reactie}.
FET usually requires typing on a confined label space (e.g., a hierarchical structure of ontology including coarse-grained types and fine-grained ones). Hence, it is time-consuming and labor-intensive to get the annotations for entity mentions of a given corpus. To mitigate the data scarcity issue, some studies~\cite{Dai2019ImprovingFE, chen2020hierarchical, li2022ultra} explored distantly-supervised FET to denoise pseudo labels incorporated from external knowledge bases. Recent studies made attempts in both few-shot~\cite{chen2020hierarchical, 10.1145/3534678.3539443} and zero-shot~\cite{chen2021empirical, ma2016label} settings where the model is provided with the label hierarchy and optional example annotations for each type. Our work focuses on the research line of zero-shot FET, where no annotation for training data is given.

\begin{table*}[!ht]
\caption{Enrichment of instance and topic information for entity types, and the corresponding generated sentences.}
\centering
\resizebox{\textwidth}{!}{
\begin{tabular}{cl}
\toprule
\multirow{3}*{Artist}&\textbf{Instance}: Leonardo Da Vinci \quad\quad \textbf{Topic information}: creativity, art history, style, etc.\\
~&\textbf{Generated Context}: The painting depicts Christ on his way to Calvary, surrounded by angels who are \\
~&carrying him up into heaven. It has been dated between 1475 and 1480. According to art historian Jos Mareda, \\
~&it shows ``the influence of \underline{Leonardo da Vinci}'', but also that ``of Giotto''. \\ 
\midrule
\multirow{3}*{Cemetery}&\textbf{Instance}: Cenotaph\quad\quad\textbf{Topic information}: tombstones, grave markers, cremation, etc.\\
~&\textbf{Generated Context}: During her visit to Paris, Emily made sure to explore the famous \underline{Père Lachaise}, where \\
~&notable figures such as Oscar Wilde and Jim Morrison were laid to rest.\\
\midrule
\multirow{3}*{Sports team}&\textbf{Instance}: The New York Yankees \quad\quad\textbf{Topic information}: training, games, rivalries, etc.\\
~&\textbf{Generated Context}: October 5, 2013: In the final game played at Yankee Stadium, \underline{the New York Yankees} defeat\\ 
~&the Oakland Athletics 7-3 behind four home runs from Alex Rodriguez and six RBI from Mark Teixeira.\\
\bottomrule
\end{tabular}
}
\label{tab:gen_context}
\vspace*{-3mm}
\end{table*}

\old{
For zero-shot FET, since the label ontology is a structural view of concepts representing relations among them~\cite{mao2020octet, shen2018entity, vrandevcic2012wikidata}, existing methods typically leverage the ontology structure to more accurately characterize the label space~\cite{geng2021ontozsl, zhang2020mzet}. 
}
\new{
An ontology is a structural view of concepts representing relations among them~\cite{mao2020octet, shen2018entity, vrandevcic2012wikidata}.
For zero-shot FET, existing methods typically leverage the ontology structure to accurately characterize the label space~\cite{geng2021ontozsl, zhang2020mzet}.}\
There are also studies that combine ontology structure with external knowledge sources such as Wikipedia label descriptions~\cite{obeidat2019description, zhou2019zero, liu2021fine}. However, these approaches often require substantial human annotations on the source domain or manually selected representative mentions.
The recent development of language models has shed new light on FET, due to their remarkable text representation power~\cite{devlin2018bert} and ability to serve as a comprehensive knowledge source~\cite{Petroni2019LanguageMA}. However, though they showcase incredible ability in general zero-shot tasks~\cite{Zhong2021AdaptingLM}, their performance on zero-shot FET~\cite{qin2023chatgpt} is not satisfactory.
This is because as the ontology layer goes deeper and the entity types become more fine-grained, language models face increasing challenges in distinguishing between these types due to their subtle semantic differences, and lower frequency of occurrence in pre-trained texts.
Hence, how to fully interpret and leverage the structure/hierarchy of the ontology can play a decisive role in zero-shot FET. 

In this paper, we present a novel zero-shot FET framework to enrich the ontology structure by incorporating instance and topic information, and design a coarse-to-fine typing algorithm that effectively utilizes the enriched information. Specifically, we first enrich the original ontology structure with two types of information in the form of words and phrases (shown in Table~\ref{tab:gen_context}): (i) \textit{instance information} gives concrete demonstrations of a specific type, which are crucial for training sample augmentation and (ii) \textit{topic information} provides type-associated key phrases to distinguish contexts of similar fine-grained types. Based on the automatically-enriched ontology, our coarse-to-fine entity typing algorithm generates pseudo-training sentences containing enriched instances, by incorporating reward and penalty mechanisms into language model decoding. Then we model entity typing as a natural language inference (NLI)~\cite{li2022ultra, komarlu2023ontotype} task, where we train the entailment model with contrasting topic information to interpret the fine-grained hierarchical semantics.
The proposed framework is evaluated on three challenging fine-grained entity typing benchmarks to verify its effectiveness.

To sum up, this study contributes to the state-of-the-art of fine-grained entity typing in the following aspects:
\begin{enumerate}[leftmargin=*]
\item We propose to enrich the existing ontology, where two types of information, \emph{instances} and \emph{topics} are leveraged to help describe and distinguish the semantics of fine-grained types.

\item  We design a zero-shot and ontology-guided entity typing framework to make good use of the enriched information. By generating instance-based pseudo-training data and modeling entity typing as an NLI task, we are able to better distinguish fine-grained entity types and encode label dependency.

\item Empirical studies verify the general effectiveness of our method on fine-grained entity typing. We also conduct comprehensive analyses for interpreting the performance of \ours.

\item We also demonstrate that \ours can work on different test sets which are free to have unseen and even more fine-grained types.
\end{enumerate}

\section{Related Work}

\paragraph{Fine-Grained Entity Typing with Supervision.}
The goal of fine-grained entity typing (FET) is to determine the type of entity given a particular context and, optionally, a label hierarchy~\cite{ling2012fine, yosef2012hyena}. Previous studies typically use pre-defined ontology hierarchy and co-occurrence structures estimated from data to enhance the models. 
To this end, \cite{ren2016afet, xu2018neural, chen2020hierarchical} design new loss functions to exploit label hierarchies in the ontology. \cite{shimaoka2016neural, murty2018hierarchical} embed labels into a high-dimension or a new space as representations. 
\cite{liu2021fine} explores label dependencies and designs label reasoning mechanisms for effective decoding. 
There are also studies that exploit the co-occurrence structures including enriching label representations by introducing associated labels \cite{xiong2019imposing}, requiring latent label representation to reconstruct the co-occurrence structure~\cite{lin2019attentive}, or limiting the label range for classification during prediction~\cite{rabinovich2017fine}. There are two problems for this research line. First, they neglect the very essential problem inherent in the given ontology, and build blocks to fit in with the potentially problematic structure, which may bring the noise. Second, the experiments are all conducted with direct or indirect supervision, which poses a great challenge and burden for data annotation. To alleviate the data scarcity issue, the setting of zero-shot fine-grained entity typing is introduced.

The most relevant work to us in this research line is LITE~\cite{li2022ultra}, which makes use of the indirect supervision from NLI to infer type information. However, LITE still requires supervision for training, which is different from our zero-shot setting. 
\old{Also, they do not consider the hierarchical structure of ontology as they perform typing, which could limit the performance.}\
\new{Also, it does not consider the hierarchical structure of ontology in typing and thus limits its performance.}

\paragraph{Zero-Shot Fine-Grained Entity Typing.}
Zero-shot fine-grained entity Typing has been explored widely to alleviate the data scarcity issue. Existing zero-shot FET frameworks infuse multiple sources of information. \citet{zhang2020mzet} focuses on capturing the relationship between unseen and seen types in order to produce label representations for the unseen types. \citet{ma2016label} enhances label representations by incorporating hierarchical and prototypical information derived from manually selected context-free entities as prototypes. \citet{yuan2018otyper} maps mention embeddings to the type embedding space by training a neural model to combine entity and context information. \citet{obeidat2019description, zhou2019zero} define unseen types by generating type embeddings from Wikipedia descriptions. \citet{chen2021empirical} fuses the combination of three external knowledge, contextual, pre-defined hierarchy, and label descriptions. Then three independent modules are trained for the knowledge, whose results are further combined to get the final results. Nonetheless, these zero-shot learning algorithms still require extensive annotations on the source domain or manually selecting high-quality representative mentions. 

OntoType~\cite{komarlu2023ontotype} is the most relevant work in a zero-shot setting compared with us, which is also an ontology-guided framework. However, OntoType is a non-training approach that has little interplay between the model and task-specific ontology. Therefore, they are poor at fine-grained type identification and their headword parsing can fail on domain-specific typing tasks. 
\definecolor{c1}{HTML}{A9AC5D}

\section{Methodology}

\begin{figure*}[h]
\centering
\includegraphics[width=1.0\textwidth]{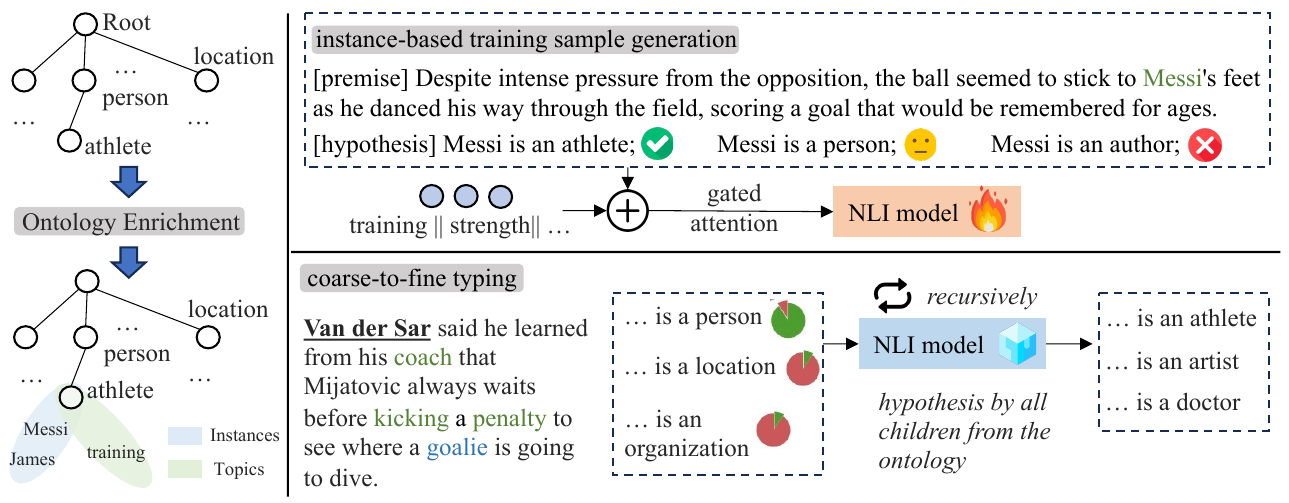}
\caption{\ours framework. The left part is the ontology enrichment for instance and topic information. Then we leverage instances to generate pseudo-training data. Together with topics, we train an NLI model, which will support coarse-to-fine typing during the inference stage.}
\label{fig:framework}
\end{figure*}

In this section, we detail the framework of \ours, which consists of two parts: (1) ontology enrichment with instance and topic information, and (2) zero-shot coarse-to-fine typing. We will first automatically enrich the given ontology structure with topic words and instances. The instances are then used to generate pseudo-training samples for the zero-shot setting, where the topics are infused to train a natural language inference (NLI) model. During the inference stage, we iteratively employ the NLI model in a coarse-to-fine manner over the ontology until we finally reach the answer. The overall architecture is shown in Figure \ref{fig:framework}.

\subsection{Automatic Ontology Enrichment}

Despite the ``is-a'' relation existing in ontology structures, we argue that enrichment in an ontology using instances and topics could really help distinguish fine-grained types. Examples of enrichment for instances and topics are displayed in Table~\ref{tab:gen_context}.

\paragraph{Enrichment for Topic Information.} 
Topic words and phrases are indicatives of the contextual understanding, and help to disambiguate between different entity types that appear similar~\cite{ajjour2023topic}. For example, entity types such as biologist and chemist may seem similar, but topic words related to each type such as \{\textit{genetics, ecology, viruses, animals}\} and \{\textit{elements, molecules, equilibrium, catalysts}\} can improve the contextual understanding and better distinguish these two types. We employ SeeTopic~\cite{zhang-etal-2022-seed}, an unsupervised framework to generate topics using fine-grained types as query words. 
Specifically, for every type node, we first select the top 20 related documents in a large Wikipedia corpus by implementing Elasticsearch~\cite{kononenko2014mining}. By only compiling documents related to the query node, we reduce noisy results and are able to extract relevant topics without intensive memory usage. Then the retrieved documents are fed into SeeTopic for topic words/phrases mining. Importantly, SeeTopic supports out-of-vocabulary search by looking for semantically similar queries via PLM embeddings, which is necessary for automating this process.

\paragraph{Enrichment for Instance Information.} 
Instances are concrete examples of entity types that are in the ontology tree structure. This level of granularity 
\old{provides for a more}\new{helps}\ detailed classification within a broader categorical structure~\cite{wu2012probase}, thus enhancing the understanding of fine-grained entity types. We leverage SECoExpan~\cite{zhang2022entity} on the Wikipedia corpus to generate instances, which are divided into the following two stages: 
(1) LM-based methods in a question-answering style~\cite{jiao-etal-2022-open} for seeds curation, and (2) SECoExpan over the generated seeds to expand instances with the Wikipedia corpus.

For seed generation, we leverage the question-answering framework and the given Wikipedia corpus. Providing the entity type $T$, we first retrieve related sentences in Wikipedia. Then for every retrieved sentence <S> we curated a question-answering template inspired by~\cite{jiao-etal-2022-open} as follows for better comprehension:

\noindent
\textit{[CLS]What is the instance of <T> in this sentence?[SEP]<S>[SEP]}

Here, \textit{[CLS]} is the special classification token, and \textit{[SEP]} is the special token for separation. $T$ denotes the entity type indicated, \textit{<S>} is the retrieved sentence in Wikipedia. An example is shown in the following:

\noindent
    \textit{[CLS] What is the instance of <movie> in this sentence? [SEP] Lepa Shandy was a Nigerian Yoruba movie that was produced by Bayowa and eventually became a very successful project. [SEP]}

In the above example, the answer would be ``Lepa Sandy''. We proceed with this process for $n$ times to get the $n$ total instance seeds for every type $T$ in the ontology structure. After obtaining instance seeds, we feed them into SECoExpan to automatically mine a sufficient amount of instances for each entity type \textit{T} based on them.
 
\subsection{Coarse-to-fine Typing}

This section presents our ontology-guided zero-shot approach \ours based on the restructured ontology. The overall architecture is shown in the right part of Figure \ref{fig:framework}. 
Building FET models typically requires training sequence labeling models that recognize entities of specific types under given contexts.
First, we construct contextualized training samples for each fine-grained type based on their context-free enriched instances. We then train an entailment model with the enriched information from topic words/phrases to distinguish the fine-grained types. During test time, we make final predictions by inferring with the entailment model in a top-down recursive manner following the ontology structure. 
\paragraph{Contextualized Training Data Construction.} 
We propose to construct contextualized training samples for each type by leveraging an LLM to generate a sentence that contains an instance $\bs{e}$ belonging to that certain type. One intuitive way for generation is to use the instance $\bs{e}$ as the starting word/phrase. However, such an approach can only create sentences with the instance being at a fixed position, while entities could appear at the beginning, middle, or end of sentences in real practice. Therefore, we propose a rewarding mechanism in LM decoding to generate diverse training samples that contain the instance $\bs{e}$.
Specifically, we use a multiplier to rescale the logits $\bs{l}_i$ of each token $x_i$ at the output softmax layer:
\begin{equation}
\label{eq:temp_prob_mod}
p_{\mathcal{G}}(x_i|\bs{x}_{<i}) = \frac{\exp(\bs{l}_i/\omega)}{\sum_{j=1}^{|V|}\exp(\bs{l}_j/\omega)}, 
\end{equation}

\begin{equation}
    \omega = \begin{cases}
\tau \alpha & x_i \in \bs{e} \land x_i \notin \bs{x}_{<i} \\
\tau \beta & x_i \in \bs{x}_{<i} \\
\tau & \text{else}
\end{cases},
\end{equation}
where $\omega > 0$ is the sampling temperature ($\omega \to 0$ approximates greedily picking the most probable next token; $\omega \to \infty$ induces a uniform distribution); $\alpha$ and $\beta$ are scaling hyperparameters of $\tau > 0$: we set $\alpha > 1$ to encourage tokens in the target entity $\bs{e}$ to have a higher probability of getting generated, and we set $\beta < 1$ to discourage the generation of repetitive tokens that already appeared in the sequence $\bs{x}_{<i}$ to mitigate degenerate repetition, a common issue in text generation. In our experiments, we 
employ CTRL~\cite{keskar2019ctrl}, a 1.6B generative language model for generation and choose ``Wikipedia'' as the generation style. 
We select the generated sentences which have a higher conditional generation probability than the average of all generated ones~\cite{meng2022generating}, and filter out samples that do not contain the given instance.

Following the architecture of natural language inference (NLI)~\cite{bowman2015large}
, we construct training samples for NLI labels with respect to the ontology structure, and contrastively train an entailment model for fine-grained type prediction. Taking the generated sentence $\bs{x}$ as the ``premise'', the ``hypothesis'' is a statement \textit{``[instance] is a [entity type]''}. If the entity type is the original one that generates the instance, the label is set as \textit{Entailment}. For \textit{Contradiction} labels, we choose the fine-grained entity type in other branches. ``Neutral'' labels are paired with coarse-grained entity types in the same branch to help better distinguish coarse-grained information from fine-grained one.

\paragraph{Training Paradigm.} 

With the generated dataset, we are now able to train an entailment model together with topics to further capture the nuances in fine-grained entity types. Specifically, we build upon the RoBERTa~\cite{liu2019roberta} model with the Transformers~\cite{vaswani2017attention} architecture. The topics are concatenated and sent to the encoder to get embedding $H^t$. Then a gated attention module is designed to incorporate the topic information (during inference, we use the keywords/phrases in the context as an approximation to topics):

\begin{equation}
    \lambda = \sigma(W_\lambda H^t+U_\lambda H^c)
\end{equation}

where $W_\lambda$ and $U_\lambda$ are learnable parameters, and $H_c$ is the context embedding of the input sentence. $H^c$ and $H^t$ are then fused for an effective representation $H = H^c+\lambda\tilde{H}$.

One challenge for training the entailment model on the synthetic dataset is that the generated samples may contain noises. 
Standard supervised training on noisy datasets may raise the risk of overfitting and degraded performance since some easy categories converge faster than hard categories.
To improve the noise-robustness of entailment model training for better generalization, we simply apply a noise-robust loss, generalized cross-entropy (GCE) loss~\cite{zhang2018generalized}, which incorporates $q$-order entropy~\cite{ferrari2010maximum} into the standard cross-entropy loss.
Specifically, the calculation of GCE loss for the entailment model $\mathcal{F}$ is as follows:

\begin{equation}
\label{eq:gce_loss}
\mathcal{L}_{\text{GCE}} = \sum_{i=1}^n \frac{1 - \mathcal{F}_{y_i}(\bs{x}_i)^{q}}{q},
\end{equation}
where $\mathcal{F}_{y_i}(\bs{x}_i)$ is the model's predicted probability of sequence $\bs{x}_i$ belonging to the corresponding entailment label $y_i$.
When $q \to 1$, $\mathcal{L}_{\text{GCE}}$ has better noise-robustness; when $q \to 0$, $\mathcal{L}_{\text{GCE}}$ approximates the standard cross-entropy loss.
\section{Experiments}

\begin{table*}[t]

\centering\centering\setlength{\tabcolsep}{2.8pt}
\small
\caption{Results on the test sets for three benchmark datasets, FIGER, OntoNotes, and BBN. We compare with baselines from two different settings, a fully/distantly-supervised setting and a zero-shot setting. The best results for each setting are highlighted in \textbf{bold}. * indicates results from GPT-3.5-turbo.}\label{table:res}
\begin{tabular}{lccccccccc}
\toprule
\multirow{2}{*}{Models} &
\multicolumn{3}{c}{FIGER} & \multicolumn{3}{c}{OntoNotes}&\multicolumn{3}{c}{BBN}\\
\cmidrule{2-4}
\cmidrule{5-7}
\cmidrule{8-10}
& Acc. & Micro-F1 & Macro-F1& Acc. & Micro-F1 & Macro-F1 & Acc. & Micro-F1 & Macro-F1\\
\midrule
\textbf{\textit{Fully / distantly-supervised Setting}}\\
\midrule
AFET~\cite{ren2016afet}& 55.3&66.4&69.3 &55.1&64.7&71.1&67.0&73.5&72.7\\
UFET~\cite{choi-etal-2018-ultra} & -&-&-&59.5&71.8&76.8&-&-&-\\
BERT-MLMET~\cite{dai2021ultra} & -&-&-&67.4&80.4&85.4&-&-&-\\
LITE~\cite{li2022ultra} & - &\textbf{83.3} &\textbf{86.7}&-&\textbf{80.9}&\textbf{86.4}&- &-&- \\
\midrule
\textbf{\textit{Zero-Shot Setting}} \\
\midrule
ProtoZET~\cite{ma-etal-2016-label} &29.6&56.4&55.1&28.1&34.5&33.7&25.1&63.1&58.2 \\
OTyper~\cite{yuan2018otyper} &47.2&67.2&69.1&31.8&36.0&39.1&29.0&48.8&54.4\\
ZOE~\cite{zhou2019zero} &\textbf{58.8}&71.3&74.8&50.7&60.8&66.9&61.8&74.9&74.6\\
DZET~\cite{obeidat2019description}&28.5&56.0&55.1&23.1&28.1&27.6&-&-&-\\
MZET~\cite{zhang2020mzet}&31.9&57.9&55.5&33.7&43.7&42.3& 29.4&68.7&60.6\\
ChatGPT*~\cite{ouyang2022training} &51.7&65.3&58.3& 27.7&37.5&32.6& 25.1&55.9&50.7\\
OntoType~\cite{komarlu2023ontotype} &49.1&67.4&75.1&65.7&73.4&81.5&-&-&- \\
\ours& 56.3&\textbf{72.7}&\textbf{78.6}&\textbf{68.6} &\textbf{76.3} &\textbf{83.4} &\textbf{68.5} &\textbf{80.1} &\textbf{81.7} \\
\bottomrule
\end{tabular}

\vspace*{-3mm}
\end{table*}

\subsection{Dataset}

In our experiments, we use a wide range of fine-grained entity typing datasets, including OntoNotes~\cite{gillick2014context}, BBN~\cite{weischedel2005bbn}, and FIGER~\cite{ling2012fine},
to verify the effectiveness of our proposed framework. The detailed statistics of the three datasets are shown in Table~\ref{table:datasets}. Note that in the zero-shot setting, we are directly inferring for the test sets.

\begin{table}
\centering
\small
  \caption{Dataset Statistics}
  \label{table:datasets}
  \begin{tabular}{lccc}
    \toprule
    Datasets&OntoNotes&BBN&FIGER\\
    \midrule
    \# of Types &89&46&113 \\
    \# of Documents &300k&2311&3.1M \\
    \# of Entity Mentions &242k&100k&2.7M\\
    \# of Train Mentions &223k&84k&2.69M\\
    \# of Test Mentions &8963&13766&563\\
    \midrule
    \# of total instances& 2377&1561&2453 \\
  \bottomrule
\end{tabular}
\end{table}

\subsection{Baseline models}

We compare our framework with baselines from two settings: (i) supervised and distantly-supervised fine-grained methods, and (ii) zero-shot fine-grained methods.

For supervised and distantly-supervised fine-grained entity typing, we have the following baselines: (1) \textbf{AFET}~\cite{ren2016afet} proposes a method for embedding both clean and noisy entity references individually. The technique leverages a defined type hierarchy to formulate loss functions and integrates them into a unified optimization problem to calculate the embeddings of references and type paths. (2) \textbf{UFET}~\cite{choi-etal-2018-ultra} predicts open types without a pre-defined label structure and is trained using a multi-objective approach that combines supervision from the headwords and prior information from entity linking in Wikipedia.
(3) \textbf{BERT-MLMET}~\cite{dai2021ultra} is a model that starts with BERT-base and fine-tunes it using supervision from headwords and entity-type hypernyms extracted from Hearst patterns. The resulting model is used to predict ultra-fine entity types and produce fine-grained entity types by means of a simple type mapping process. (4) \textbf{LITE}~\cite{li2022ultra} borrows indirect supervision from NLI to perform entity typing. It also involves a type-ranking module to help with generalizing prediction with disjoint type sets.

For zero-shot fine-grained entity typing baselines, we have the following baselines: (1)~\textbf{OTyper}~\cite{yuan2018otyper} is a neural network trained on a limited training dataset in shallow levels of ontology. This model is then evaluated on the Open Named Entity Typing task, which is actually zero-shot since the fine-grained target types are not known in advance. (2)~\textbf{ZOE}~\cite{zhou2019zero} uses a new type taxonomy defined as Boolean functions of Freebase types and determines the type of a given entity reference by linking it to the type-compatible Wikipedia entries.
(3)~\textbf{DZET}~\cite{obeidat2019description} leverages the type descriptions available from Wikipedia to build a distributed semantic representation of the types and aligns the target entity mention and their corresponding type representations onto the known types.
(4)~\textbf{MZET}~\cite{zhang2020mzet} leverages the semantic meaning and the hierarchical structure into the type representation. With a memory component to model the relationship between the entity mentions and types, the method transfers the knowledge from seen types to label the unseen types. (5) \textbf{ChatGPT}~\cite{ouyang2022training} is a generative large language model aiming for generic AI applications. Pre-trained on the large-scale corpus with reinforcement learning techniques, ChatGPT provides a richer knowledge source for entity typing. (6) \textbf{OntoType}~\cite{komarlu2023ontotype} is an ontology-guided framework that leverages the weak supervision of pre-trained language models and headwords, which are further used to match the fine-grained types to type ontology.

\subsection{Experiment details}
For each type in the ontology structure, we enrich $30$ instances and $5$ topics. The number of enriched topics directly adopts the hyperparameter choices from the original topic discovery paper~\cite{zhang-etal-2022-seed}, which finds that the quality of top-$5$ terms is mostly reliable and contains less noise. For the number of instances, we experiment with 10, 20, 30, 40, and 50, with the detailed performance shown Table~\ref{table:instance_num}. Setting a number equaling 30 achieves the best results. Potential reasons are (i) we need a sufficient number of pseudo samples in training, and 30 ensures a good number of samples, (ii) too many samples (e.g., 40 or 50 instances) could increase noises (e.g., intrusion between some categories). Also, generating top-40 and top-50 instances per type could take a longer time for instance generation. Note that each type does not always have $30$ instances, which is due to the characteristics of each type. For example, ``railway'' and ``time'' do not have 30 instances. The total number of instances enriched and the original number of nodes are shown in Table~\ref{table:datasets}. When generating pseudo-training data, we generate one sample for each instance. When training the entailment model, we set the number of epochs to $10$, and the learning rate to $1e-5$. It takes $1$ hour to run our method with 2 Nvidia A6000 GPUs, and around 20 minutes to do inference for OntoNotes. Note that the inference time varies with respect to different numbers of test samples.

\begin{table}[!t]
\centering\setlength{\tabcolsep}{6.8pt}
\small
  \caption{Experiment results using different number of instances on the test set of OntoNotes dataset.}
  \label{table:instance_num}
  \begin{tabular}{lccccc}
    \toprule
    \# instances&10&20&30&40&50\\
    \midrule
    Accuracy & 60.9	&64.1	&68.6&	67.4&	65.9 \\
    Micro-F1 & 71.0	&73.5	&76.3	&74.9&	73.6 \\
    Macro-F1& 76.4&	79.2	&83.4&	81.8	&79.8 \\
  \bottomrule
\end{tabular}
\end{table}



\subsection{Results}

Table~\ref{table:res} presents the performance of all methods on the test set of three benchmark datasets. All the best results are shown in bold. Based on the results, we have the following observations:

(1) \ours achieves superior performance on almost all the datasets with dramatic improvements compared with zero-shot baselines. Specifically, \ours achieves an absolute improvement of $+2.9$, $+2.9$, $+1.9$ on Accuracy, Micro-F1, and Macro-F1 respectively on the OntoNotes dataset, compared with the previous state-of-the-art zero-shot entity typing model $OntoType$. For the BBN dataset, \ours also achieves a consistent performance with a performance boost of $+7.7$, $+5.4$, and $+6.8$ accordingly compared with $ZOE$. For the FIGER dataset, we fall a little behind $ZOE$ on Accuracy and Micro-F1 scores. Nevertheless, we achieved a $+3.5$ gain on the Macro-F1 score. We also observe that \ours is competitive compared with supervised or distantly-supervised methods.

(2) We find that the improvement in metric Macro-F1 is much better than that of Micro-F1. Macro-F1 weights every class equally by first calculating $P_\text{Ma}$ and $R_\text{Ma}$ and then taking the average over classes. Conversely, Micro-F1 weights every sample equally. The results indicate that \ours not only boosts the performance for the majority class labels such as ``/person'' and ``/organization'', but also further enhances the typing accuracy on minority class labels such as ``/event/war'' by a large margin.

(3) We also investigate the performance against a recent LLM, ChatGPT, by evaluating fine-grained entity typing through OpenAI API ``gpt-3.5-turbo''. Specifically, we use the instruction ``\textit{Give the fine-grained entity types of the given entity mentioned in the sentences below. Be concise and you can ONLY use types from this list of {possible entity types}.[sentence] \{sentence\} [entity mention] \{entity mention\}}''. We give 3 demonstrations to help with typing. As shown in Table~\ref{table:res}, ChatGPT, though deemed promising, proved unsatisfactory in fine-grained entity typing. The reasons are twofold. Firstly, LLMs such as ChatGPT do not have the background of type structure, i.e., ontology, and instead tend to generate free-form entity types, which could be noisy. Also, though we provide strict demonstrations as templates, LLMs tend to generate redundant sentences as output, which limits their accuracy.

(4) We are curious about the phenomenon as to why \ours surpasses ZOE by a large margin on OntoNotes and BBN datasets, but falls a little behind on the FIGER dataset. One of the reasons is that there are many nested entity typing problems existing in FIGER dataset. Consider the following example: ``Nine of the individuals elected are \textbf{UW} faculty members.'' ZOE directly grounds \textbf{UW} in the corresponding Wikipedia entries and will obtain ``/organization/educational\_institution'' almost surely. However, in \ours, we leverage prompting predictions as a major source, which may lead to a contextualized understanding of ``UW faculty members'' and type ``UW'' with coarse-grained type ``/person''. We provide a more detailed error analysis in Section~\ref{sec:error_analysis}.
\section{Analysis}

\begin{table}[!t]
\centering
\small
  \caption{Ablation studies for different components in \ours. The Accuracy and F1 scores are evaluated with the test set of the OntoNotes dataset.}
  \label{table:ablation}
  \begin{tabular}{lccc}
    \toprule
    Model&Acc.&Micro-F1&Macro-F1\\
    \midrule
    \ours & 68.6&76.3&83.4 \\
    \midrule
    \quad w/o topics & 67.0&74.1 &81.9 \\
    \quad w/ 3 instances & 59.8 & 70.5 & 75.6 \\
    \quad w/o coarse-to-fine & 67.5&75.7&82.1\\
    \quad w/o GCE loss & 66.4&74.5&81.6\\

  \bottomrule
\end{tabular}
\end{table}

\subsection{Ablation Study}

To dive into the effectiveness of different components in \ours, we conduct a comprehensive analysis on the test set of OntoNotes. Table~\ref{table:ablation} summarizes the results.

\paragraph{Effect of topical enrichment.} We explored how performance changes when topics are removed. As shown in Table~\ref{table:ablation}, the performance decreases when we use generated plain text for NLI training, and do not take keywords/phrases into consideration during inference. This result attests to the notion that topical enrichment contributes significantly to understanding and resolving complex semantic relationships in NLI tasks.

\paragraph{Effect of enrichment for instances.}
We also explored how instance enrichment influences performance.  Notably, the training data are constructed from instance information, without which the framework could not be complete as in a zero-shot setting. Therefore, we conducted experiments where only 3 instances were used for training data construction, to approximate the ablation results. The enriched ablation results are shown in Table~\ref{table:ablation}. This indicates that instances also contribute to the final performance.

\paragraph{Effect of inference style.} To investigate how coarse-to-fine inference on ontology contributes to the final performance, we conduct another experiment where we treat all the nodes in the ontology at one plain level.
We notice that the performance does not drop as much as we expect, since the traditional NLI model may give more weight to safe answers as coarse-grained types. The potential reason could be our contrastive label design when training the NLI model, which improves the model's ability in capturing fine-grained information.

\paragraph{Effect of loss function design.}

We also tested how the design of GCE loss affects the performance. In this experiment, we replace the GCE loss with traditional cross-entropy loss and everything else remains the same. As shown in Table~\ref{table:ablation}, replacing GCE loss leads to worse performance, and even less accuracy than removing topics. This confirms the necessity of using a noise-tolerant training setting. The reason is that GCE loss can prevent the model from overfitting to false negative labels, thus improving the performance.

\subsection{Transferibility Test}

To demonstrate \ours's ability in entity typing to unseen and even more fine-grained samples, we employ the UFET benchmark dataset~\cite{choi-etal-2018-ultra} for comparison. The UFET dataset consists of two parts, human-labeled data with 5994 instances split into train/dev/test by 1:1:1, and distant supervision data including 5.2M instances that are automatically labeled by external knowledge bases. There are a total of 10,331 ultra-fine-grained types that are not organized in an ontology. We focus on the human-labeled data with 1,998 test samples in a zero-shot setting, and follow the original design to evaluate loose macro-averaged precision (P), recall (R), and F1 score. 

``Direct NLI'' refers to the baseline that is pre-trained on MNLI, and then predict directly without any tuning on the test set. We also introduce the previous state-of-the-art (distantly) supervised models as references. We introduce two ways to apply \ours to the UFET dataset. One is to directly employ the trained NLI model with FET datasets to UFET datasets (direct \ours), the other is to first generate pseudo training data for each type\footnote{Since the label space is large, we generate 5 training samples for each entity type.} and then train a new NLI model with these data (\ours).

\begin{table}[!t]
\centering
\small
  \caption{Results on the ultra-fine entity typing task.}
  \label{table:ultra-fine}
  \begin{tabular}{lccc}
    \toprule
    Model&P&R&F1\\
    \midrule
    \textbf{\textit{supervised setting}} \\
    MLMET~\cite{dai2021ultra} & \textbf{53.6}&45.3&49.1 \\
    LITE~\cite{li2022ultra}&52.4&48.8&\textbf{50.6} \\
    \midrule
    \textbf{\textit{zero-shot setting}} \\
    direct NLI& 1.5&7.1 &2.5 \\

direct \ours & 7.2 & 17.5& 10.2\\
\midrule
\ours &31.4&\textbf{53.1}&39.5 \\
    
  \bottomrule
\end{tabular}
\end{table}

The results are shown in Table~\ref{table:ultra-fine}. We observe that directly employing the previously trained NLI model already achieves competitive performance against the baseline model NLI, which verifies \ours's ability to distinguish fine-grained types. There is still a gap between previous supervised state-of-the-art methods due to the difficulty and massive unseen, ultra-fine-grained types. Additionally, we observe a massive performance increase when $5$ instances (training samples) are generated for each type in UltraFine. Although we are still around $10$ points of F1 behind the state-of-the-art model, we achieve the best performance of Recall so far. This is due to the characteristic of NLI models, which tend to give more answers that they think are suitable. Despite this, \ours still demonstrates its superiority considering that this is actually a $5$-shot training compared with full fine-tuning of state-of-the-art models.

\subsection{Error Analysis}\label{sec:error_analysis}

To shed light on how to further improve \ours, we examine the bottleneck of the method by conducting an error analysis here. Specifically, we randomly sampled $100$ erroneous cases predicted by \ours from OntoNotes test set, and list three major categories of errors as follows.
For each example, we bold the entity mention, and provide the annotations from datasets in \textit{\textcolor{blue}{blue}}.

\smallskip
\noindent
\textbf{1: Nested entity spans.}
\ours sometimes assigns mistakenly the types of the entire entity to that of its nested entity as shown below. About $27\%$ of the errors are partly due to the nested issues.

\noindent
\textbf{Sentence:} 
\emph{The \textbf{UAW} \emph{\textcolor{blue}{[/organization/company]}} leaders are trying to silence dissidents who charge the union are too passive in the face of GM layoffs.}   

Here, \ours incorrectly types ``UAW'' as ``/person/leader''. This can be attributed to the contextual understanding of the sentence, which tends to treat the entire entity span ``UAW leaders'' as a mention.

\smallskip
\noindent
\textbf{2: Incorrect fine-grained inference.}
Even though \ours does generally well in fine-grained entity typing, it could be distracted when there is not enough information given. This is the major reason behind $24\%$ of the errors.

\noindent
\textbf{Sentence:} 
\emph{Harry has avoided all that by living in \textbf{a Long Island suburb}\emph{\textcolor{blue}{[/location/city]}} with his wife, who's so addicted to soap operas and mystery novels she barely seems to notice...}  

Here \ours predicts the entity type as ``/location'' without going further fine-grained. A potential reason is the lack of knowledge of the phrase ``Long Island''.

\smallskip
\noindent
\textbf{3: Debatable predictions against the ground truth.}
\ours sometimes disagrees with what is shown as the ground truth but it is arguable which one is a better answer. This type of error contributes another $9\%$.

\noindent
\textbf{Sentence:} 
\emph{Currently, it has an annual production capacity of \textbf{4,000 tons} \emph{\textcolor{blue}{[/other/food]}}.}   

Here \ours predicts the entity is a ``/product'', which is a supertype of ``food'' in our refined ontology. 
We believe that ours is a better answer since without additional context no one can assert it is food.

\section{Conclusion}

In this project, we study the problem of zero-shot fine-grained entity typing (FET) via our designed framework \ours which leverages the automatically-enriched instances and topic information of ontology. We first generate pseudo-training data with instances with reward mechanisms to encourage diversity in the generation. Then we incorporate topic information in self-attention to better align contextualized information. We also use the generalized cross-entropy loss to allow noise in the generated training sentences. Experiments on three benchmark datasets demonstrate that our method outperforms previous zero-shot baselines. Additionally, we also apply \ours to unseen and even fine-grained entity types to verify the transferability.

\section*{Limitations}
Our work on \ours is subject to multiple limitations. The first limitation is the time cost during inference. Since we are training an NLI model, we have to go over all the entity types in the label space, the inference time could be very long for datasets with massive labels, such as UltraFine. One potential way to accelerate this process is to employ a label-binding cross-encoder framework~\cite{pang-etal-2020-fastmatch}. Secondly, we only explored the assertive statement as the hypothesis in our training data construction, future work may introduce more diverse sentences that contain more linguistic features and contextual information to further boost the performance. Finally, although we adopt topic and instance mining frameworks with good quality and performance, the pipelined approach could still introduce potential biases. Therefore, future work could target an end-to-end method.

\bibliography{anthology,custom}
\bibliographystyle{acl_natbib}

\appendix

\begin{figure*}[h]
\centering
\includegraphics[width=1.0\textwidth]{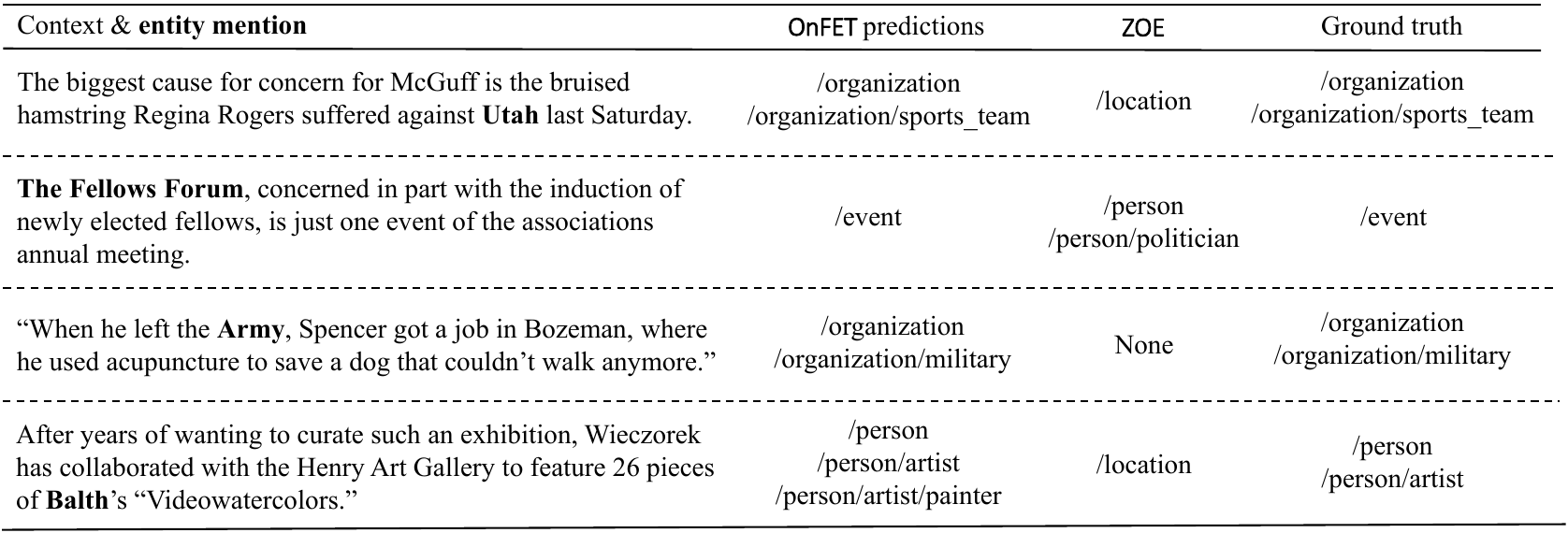}
\caption{Case study of \ours with baseline model ZOE and ground truth.}
\label{fig:case_study}
\end{figure*}

\section{Evaluation Metrics}\label{expr:metric}

Following~\cite{ling2012fine}, we evaluate our framework and compare it with baseline models using three metrics: Strict Accuracy (Acc), Micro-F1 (Mi-F1), and Macro-F1 (Ma-F1). Given a set of entity mentions $M$, ground truths and predicted types are denoted as $t_M$ and $\tilde{t}_M$ respectively. We have the following metrics:
\paragraph{Strict Accuracy} The prediction is considered correct if and only if $t_M=\tilde{t}_M$. 
$$Acc=\frac{\sum_{m\in M\sigma(t_m=\tilde{t}_m)}}{|M|}$$
\paragraph{Macro-F1} Macro-F1 is calculated using Macro-Precision ($P_\text{Ma}$) and Macro-Recall ($R_{\text{Ma}}$) where
$$P_\text{Ma}=\frac{1}{|M|}\sum_{m\in M}\frac{|t_m\cap\tilde{t}_m|}{\tilde{t}_m}$$
$$R_\text{Ma}=\frac{1}{|M|}\sum_{m\in M}\frac{|t_m\cap\tilde{t}_m|}{t_m}$$
\paragraph{Micro-F1} Micro-F1 is calculated using Micro-Precision ($P_\text{Mi}$) and Micro-Recall ($R_\text{Mi}$) where
$$P_\text{Mi}=\frac{\sum_{m\in M}|t_m\cap\tilde{t}_m|}{\sum_{m\in M}\tilde{t}_m}$$
$$R_\text{Mi}=\frac{\sum_{m\in M}|t_m\cap\tilde{t}_m|}{\sum_{m\in M}t_m}$$

\section{Case Study}
We further conduct a case study compared to some baseline models for an intuitive view as shown in Figure~\ref{fig:case_study}. Since ZOE performs the best as to the accuracy in the FIGER dataset, here we randomly grabbed several samples and compare the results generated by \ours, ZOE, and the ground truth on the FIGER dataset. We observe that \ours performs better than ZOE on these samples. Specifically, \ours well interprets the contextualized information. Consider the first example, ZOE maps ``Utah'' directly to Wikipedia entries, and obtains ``location'', whereas in the specific context, ``Utah'' actually means a sports team. In the fourth example, \ours takes advantage of the key phrase ``Videowatercolors'' and infers that ``Balth'' here refers to a painter, which is even finer than the ground truth annotation.

\end{document}